\documentclass[sigconf]{acmart}
\usepackage{amsmath}
\AtBeginDocument{%
  \providecommand\BibTeX{{%
    \normalfont B\kern-0.5em{\scshape i\kern-0.25em b}\kern-0.8em\TeX}}}

\usepackage{amsfonts}
\usepackage{amsmath}
\usepackage{algorithm, algorithmic}
\usepackage{multirow}
\usepackage{adjustbox}
\usepackage{graphicx}

\newcommand{\best}[1]{\textcolor[RGB]{205,83,76}{#1}}
\newcommand{\secondb}[1]{\textcolor[RGB]{0,115,194}{#1}}
\newcommand{\cmark}{\checkmark}%
\newcommand{\xmark}{ }%

\usepackage{amsmath}

\copyrightyear{2023}
\acmYear{2023}
\setcopyright{acmlicensed}
\acmConference[CIKM '23]{Proceedings of the 32nd ACM International Conference on Information and Knowledge Management}{October 21--25, 2023}{Birmingham, United Kingdom} 
\acmBooktitle{Proceedings of the 32nd ACM International Conference on Information and Knowledge Management (CIKM '23), October 21--25, 2023, Birmingham, United Kingdom}
\acmPrice{15.00}
\acmDOI{10.1145/3583780.3614818} 
\acmISBN{979-8-4007-0124-5/23/10}
\begin{document}

\title{Community-Based Hierarchical Positive-Unlabeled (PU) Model Fusion for Chronic Disease Prediction}


\author{Yang Wu}

\orcid{0009-0006-8438-3505}
\affiliation{%
  \institution{Worcester Polytechnic Institute}
  \city{Worcester}
  \state{MA}
  \country{USA}
}
\email{ywu19@wpi.edu}

\author{Xurui Li}
\orcid{0000-0003-1904-222X}
\affiliation{%
  \institution{Alibaba Group}
  \city{Hangzhou}
  \state{Zhejiang}
  \country{China}}
\email{xurui.lxr@alibaba-inc.com}

\author{Xuhong Zhang}
\orcid{0000-0001-7563-9915}
\affiliation{%
  \institution{Indiana University, Bloomington}
  \city{Bloomington}
  \state{Indiana}
  \country{USA}}
\email{zhangxuh@indiana.edu}

\author{Yangyang Kang}
\orcid{0000-0002-8537-0208}
\affiliation{%
  \institution{Alibaba Group}
  \city{Hangzhou}
  \state{Zhejiang}
  \country{China}}
\email{yangyang.kangyy@alibaba-inc.com}

\author{Changlong Sun}
\orcid{0009-0008-2405-3242}
\affiliation{%
  \institution{Alibaba Group}
  \city{Hangzhou}
  \state{Zhejiang}
  \country{China}}
\email{changlong.scl@taobao.com}

\author{Xiaozhong Liu}
\authornote{Corresponding author}
\orcid{0000-0003-3477-8323}
\affiliation{%
  \institution{Worcester Polytechnic Institute}
  \city{Worcester}
  \state{MA}
  \country{USA}}
\email{xliu14@wpi.edu}
\renewcommand{\shortauthors}{Yang Wu, et al.}





\begin{abstract}
  Positive-Unlabeled (PU) Learning is a challenge presented by binary classification problems where there is an abundance of unlabeled data along with a small number of positive data instances, which can be used to address chronic disease screening problem. State-of-the-art PU learning methods have resulted in the development of various risk estimators, yet they neglect the differences among distinct populations. To address this issue, we present a novel Positive-Unlabeled Learning Tree (PUtree) algorithm. PUtree is designed to take into account communities such as different age or income brackets, in tasks of chronic disease prediction. We propose a novel approach for binary decision-making, which hierarchically builds community-based PU models and then aggregates their deliverables. Our method can explicate each PU model on the tree for the optimized non-leaf PU node splitting. Furthermore, a mask-recovery data augmentation strategy enables sufficient training of the model in individual communities. Additionally, the proposed approach includes an adversarial PU risk estimator to capture hierarchical PU-relationships, and a model fusion network that integrates data from each tree path, resulting in robust binary classification results. We demonstrate the superior performance of PUtree as well as its variants on two benchmarks and a new diabetes-prediction dataset. 

\end{abstract}




\maketitle

\section{Introduction}
In the context of chronic disease detection, the collection of incomplete positive patient data, accompanied by an absence of negative samples in real-world repositories, poses a formidable challenge. This is due to the costly annotation process and the unavailability of negative data. For instance, functions like diabetes screenings may be able to label patients as ``\textit{positive}'' cases, but those who have not taken the test cannot be identified as ``\textit{negative}'' instances. Instead, they form a mixed population of both positive and negative outcomes. This issue was initially addressed by positive-unlabeled (PU) learning (\cite{denis1998pac}) and has since given rise to numerous solutions for chronic disease detection \cite{chen2020self}.

The current PU Learning techniques can be split into two distinct categories: Sample-Selection and Cost-Sensitive approaches. The former seeks to identify reliable negative examples from unlabeled data, thereafter building a supervised learning model utilizing both the positive and reliable negative data \cite{liu2003building,liu2002partially,yu2004pebl}. The spy technique \cite{liu2002partially} employs positive instances, known as "spies," to enhance the accuracy of negative sample identification. By training a classifier with these labeled positives and unlabeled data, the technique assigns probabilities to unlabeled samples, aiding in the selection of reliable negatives based on discerning thresholds. A recent study \cite{luo2021pulns} on this track used reinforcement learning to train a negative sample selector to consistently select the reliable negative samples. The other approach is cost-sensitive PU learning, which was first introduced by uPU \cite{du2014analysis}, which reweights the importance of positive and negative risks to optimize a risk estimator for PU learning. However, the flexible deep neural network used in uPU may cause the overfitting issue. To overcome this problem, \cite{kiryo2017positive} propose non-negative PU learning (nnPU) to control the training loss to be non-negative. nnPU is the current state-of-art PU learning method, and many efficient algorithms are built upon nnPU using deep neural network\cite{chen2020self,su2021positive,zhao2022dist}. 

\begin{figure}[t]
  \centering
  \includegraphics[width=\linewidth]{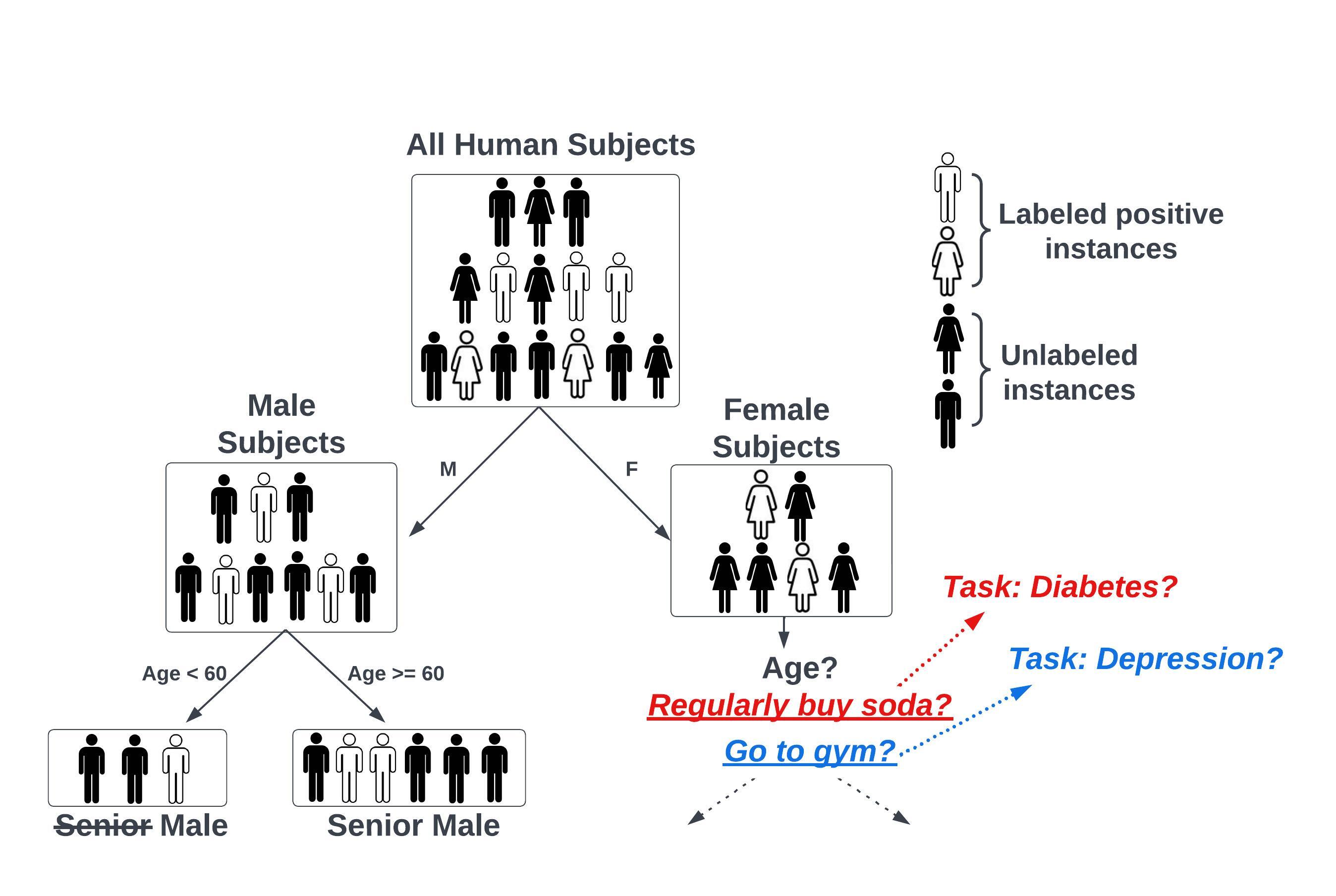}
  \caption{A toy example of diabetes diagnosis for PUtree algorithm. PUtree uses PUTreeLIME to select optimal splitting criteria, and then builds PU model on each community. Based on instances' landing node, PUtree aggregates information on all the PU models on the path to make prediction.}
  \Description{The framework of our proposed PU path fusion network.}
  \label{toy_example}
\end{figure}

Despite existing PU learning models, the needs of chronic disease detection may not be entirely met. Studies have revealed that patients across different communities can display significantly varied profiles, necessitating models which can effectively distinguish among multiple demographics, vulnerabilities or behaviors \cite{berkowitz2013clinical, vanstone2013diet}. To overcome this issue, we introduce PUtree, a novel model as shown in Figure \ref{toy_example}, which is capable of characterizing PU models corresponding to each algorithm-detected community in a hierarchical structure. Unlike conventional methods, the proposed PUtree incorporates interdependence of multiple PU models. Rather than the nodes representing independent communities, those which are siblings to one another have exclusive model parameters. This advancement is further amplified by the inheritance of information on the generational level. By being the beneficiaries of this accumulated knowledge, descendant nodes are equipped with the ancestor information to generate an enhanced deliverable via the melding of PU models. Notably, the PUtree is dynamically generated in terms of the goal of the downstream learning task. As demonstrated in Figure \ref{toy_example}, different tree structures are created when detecting diabetes and depression. This indicates that the PUtree is highly effective in fitting various objectives. The proposed Positive-Unlabeled Tree Local Interpretable Model-Agnostic
Explanations (PUTreeLIME) can be used to interpret each PU model in a meaningful way, allowing knowledge to be gathered from the data structure that is tailored to the stated task. For example, ``\textit{regularly buy soda}" might be an indicator of diabetes, and ``\textit{go to the gym}" might have greater utility when investigating depression detection. Thus, the tree can be adjusted, as necessary, to extract greater insight from the data.

Moreover, as the PUtree continues to expand, the population size of sub-communities gradually decreases, presenting the challenge of limited training resources, particularly in the case of the already scarce positive instances of chronic disease patients across all populations. To tackle this problem, a novel cross-model mask-recovery strategy is introduced to handle the sub-communities scarcity problem from data-augmentation aspect. It enables effective training of the community-based PU model on PUtree, even for small communities. For instance, when predicting diabetes within a specific community like ``\textit{non-senior, male}" we can strategically mask crucial features such as "regularly purchasing soda" based on the parent node ``\textit{male}" community PU model, while utilizing the remaining features to recover the masked information, thus achieving the goal of data augmentation.


To further leverage the rich information contained in the parent communities, we designed a hierarchical architecture for tackling the sub-communities scarcity problem from model-enhancement aspect. The proposed hierarchical PU model fusion method enhanced by an innovative adversarial risk estimator enables us to accurately assess the predictive capabilities of the model across varying community demographics. For instance, when predicting outcomes for a ``\textit{non-senior male}" individual, our ``\textit{non-senior, male}" PU model should yield predictions that are similar/cognate to the ``\textit{male}" PU model, while significantly differing from the ``\textit{senior, female}" PU model. To provide a comprehensive prediction for this non-senior male individual, we perform model fusion using the tree path (e.g., ``\textit{all human}" $\rightarrow$ ``\textit{male}" $\rightarrow$ ``\textit{non-senior male}" $\rightarrow$ ...) to generate an enhanced prediction. By incorporating this interactive framework among different community PU models, our PUtree model demonstrates superior robustness and performance.



The contributions of our work can be a threefold:
\begin{itemize}
\item  In our work, we introduce PUtree, a unique community-based PU learning algorithm that constructs distinct PU models for different communities, taking into account the variations among populations. Unlike existing methods, PUtree incorporates community interactions, allowing for a comprehensive understanding of the relationships and dependencies among diverse populations. This integration enables more accurate and contextually informed PU learning, enhancing the overall performance of the PUtree algorithm.
\item  We further introduce the novel PUTreeLIME, which enhances the interpretability and usability of PUtree by incorporating sibling community interactions to determine optimal split criteria. To mitigate the low resource problem in sub-community PU building, we employ a sophisticated mask-recovery data augmentation strategy and an advanced adversarial PU risk estimator. Additionally, our hierarchical PU model fusion captures the intricate hierarchical relationships among distinct community PU models, enabling more accurate and comprehensive predictions in community-based PU learning.
\item   We validate the effectiveness of the proposed model through comprehensive testing on a new, real-world evidence dataset that contains an abundance of features and encompasses a modest diabetic patient population. Additionally, we evaluate the model's generalizability through its performance on two popular benchmark PU learning datasets.
\end{itemize}

\section{RELATED WORK}

\subsection{Positive-Unlabeled (PU) Learning}
PU learning has been used to address different problems, including recent efforts on chronic disease prediction \cite{chen2020self}, fake news detection \cite{liu2020fned}, and recommendation system \cite{zhou2021pure}, etc. The current state-of-art methods (e.g. uPU, nnPU \cite{kiryo2017positive}) in PU learning use unbiased risk estimator \cite{du2015convex} to adjust the bias over positive and unlabeled data. Predictive adversarial learning were introduced by \cite{wei2018positive, hu2021predictive}. They used generator to produce data to fool a discriminator which can determine the generated data is positive or not. Dist-PU \cite{zhao2022dist} focuses on achieving label distribution consistency and enhancing separability between positive and negative distributions by minimizing entropy. Imbalanced PU \cite{su2021positive} is an variant of nnPU, which can learn binary classifier on imbalanced data. Imbalanced PU optimize the PU learning classifier by assuming that oversampling can tackle the imbalance issue if both P and N are available. Based on this assumption, Imbalanced PU proposes a re-weighting strategy to control the contribution of unlabeled data to conduct PU learning in a balanced manner. GenPU \cite{hou2017generative} introduces the GAN network into PU learning. The generated data is used to train binary classifier. PAN\cite{hu2021predictive} proposes a new objective based on KL-distance, and optimize the architecture of GAN. In this work, we proposed a PUtree algorithm which splits PU instances into hierarchical communities, and then a PU path fusion network is deployed to aggregate different level community information, which can be vital for various PU tasks, especially for chronic disease prediction. 


Most existing PU learning methods operate under the assumption of a known class prior, which represents the proportion of positive instances in unlabeled data. However, the class prior is often unavailable in real-world data. For instance, when detecting fraudulent transactions in financial data, the exact proportion of fraudulent transactions is typically not known in advance. To address this challenge, researchers have made significant efforts to develop techniques for estimating the class prior. For example, \cite{bekker2018estimating} deploys decision tree induction on subdomains of the data to find the lower bound of the real positive class prior. CAPU \cite{chang2021positive} jointly learns a classifier and estimates the class prior. VPU \cite{chen2020variational} tries to avoid using the positive class prior in the training process. It proposes a variational principle with Mixup regularization. In contrast to previous studies, our work tackles a more challenging task of estimating the class prior for specific communities (e.g., diabetes rate for senior males and people who regularly consume alcohol) rather than the overall population. To accomplish this, our mask-recovery strategy leverages data augmentation techniques to enhance the accuracy of class prior estimation.

\subsection{Explainable Artificial Intelligence}
Machine learning has showcased impressive performance across diverse domains, including intrusion detection, recommendation systems, and natural language processing. As the predictive capabilities of machine learning and deep learning models advance, the need to understand the factors influencing model decision-making has become more prominent. This demand has led to the emergence of Explainable Artificial Intelligence (XAI), which focuses on developing techniques and methodologies for providing transparent and interpretable explanations of model outputs. 

Existing XAI encompasses two main approaches. The first approach involves designing inherently interpretable algorithms, such as decision trees, Bayesian Rule Lists (BRL) \cite{letham2015interpretable}, and Sparse Linear Models (SLIM) \cite{ustun2016supersparse}. The second approach, known as the post-hoc class of XAI, focuses on building complex black-box models with high performance and then using reverse engineering techniques to provide explanations. One notable example of this approach is the Local Interpretable Model-Agnostic Explanation (LIME) \cite{ribeiro2016should}. LIME provides local explanations for black-box models based on the neighborhood of a specific instance of interest. In addition to local interpretability methods, there are also research works focused on achieving global interpretability. For instance, global model interpretation via recursive partitioning (GIRP) \cite{yang2018global} constructs an interpretation tree that applies to various machine learning methods, providing a global understanding of the model's behavior. In contrast to existing interpretation methods, our proposed PUTreeLIME approach takes into account the interaction among different community models within our PUtree. By considering this interaction, we are able to determine optimal split criteria for our PUtree, leading to improved overall performance and robustness. This novel approach enhances our understanding of the PUtree model and its decision-making process.

\subsection{Enhanced Learning through Augmentation and Fusion}
Data augmentation (DA) has gained significant attention in machine learning, driven by research in low-resource domains, emerging tasks, and the widespread use of large-scale neural networks that demand extensive training data. Data augmentation techniques were initially explored in the field of computer vision, where operations such as image flipping, cropping, and rotation were applied. For instance, \cite{inoue2018data} introduced a method of data augmentation by mixing images through averaging pixel values across RGB channels. Moreover, data augmentation has also emerged as a captivating avenue of exploration within the realm of natural language processing (NLP). Notable endeavors include Backtranslation \cite{sennrich2015improving}, which entails translating a sequence into a foreign language and then back into its original language. Moreover, \cite{anaby2020not} have leveraged a label-conditioned generator, fine-tuning the influential GPT-2 \cite{radford2019language} model on the training data, to generate augmented examples tailored to specific classes. In this study, we propose a data augmentation approach where we mask the top 15\% most important features as labels. We then deploy a multi-label regression model \cite{xu2013multi} to recover these masked features using the remaining features. The proposed mask-recovery data augmentation strategy effectively addresses the low-resource problem encountered as the PUtree expands to deeper levels, thereby significantly enhancing the performance of our PUtree model.

Model fusion serves as another powerful enhancement method, overcoming limitations faced by conventional machine learning techniques when dealing with complex datasets characterized by noise, imbalanced distributions, and other challenging factors. Bagging and boosting are well-established methods in traditional model fusion. A notable example of the bagging approach is the renowned random forest algorithm \cite{breiman2001random}, which constructs an ensemble of decision trees and combines their predictions through a voting mechanism. Moreover, the fusion of diverse deep learning models has exhibited impressive efficacy, exemplified by the groundbreaking fusion of heterogeneous deep networks \cite{tabik2020mnist}, which leverages the diversity of disparate models to achieve superior generalization capabilities. In our work, we employ a model fusion strategy along each distinct path of our PUtree model to generate comprehensive predictions for individual instances based on their corresponding landing nodes. This fusion framework endows our PUtree model with enhanced robustness and stability, leading to improved performance.

\section{METHODOLOGY}
In this section, we begin by introducing the PU problem definition in Section 3.1. We then proceed to present the overall PUtree algorithm in Algorithm \ref{alg:1} and Figure \ref{model_overview} in Section 3.2. Next, we demonstrate the process of selecting optimal split criteria, constructing PU models for each child-node, and hierarchical path fusion framework in detail in Sections 3.3, 3.4, and 3.5, respectively.

\subsection{Problem Definition}
Assume there is an underlying distribution $P(X,Y)$, where $X \in \mathbb{R}^{d}$ and $Y \in \{ \pm 1 \}$ are the input and output variables, respectively. PU learning assumes that labeled positive instances P are sampled independently from P marginal $\mathcal{P}_{p}(x)=\mathcal{P}(x|Y=+1)$, and the unlabeled instances U are sampled independently from mixed marginal $\mathcal{P}(x)=\pi \mathcal{P}_{p}(x)+(1-\pi)\mathcal{P}_{n}(x)$ where $\pi$ is the positive class prior and $\mathcal{P}_{n}(x)=\mathcal{P}(x|Y=-1)$ is the N marginal. \cite{du2015convex} proposed that minimizing the risk estimator can be used as the objective function in positive-unlabeled learning, i.e.
\begin{equation}
    \begin{aligned}
        R_{pu}(g) = &\pi \mathbb{E}_{p(x|Y=+1)}[l(g(x),+1)] + \\
        &(\mathbb{E}_{p(x)}[l(g(x),-1)] - \\
        &\pi\mathbb{E}_{p(x|Y=+1)}[l(g(x),-1)])
    \label{upu}
    \end{aligned}
\end{equation}

where $l(\cdot, \cdot)$ is any surrogate loss of zero-one loss \cite{du2015convex}. According to Eq.\ref{upu}, an unbiased PU learning (uPU) \cite{du2015convex} can be optimized based on Eq.\ref{upu2}.
\begin{equation}
    \begin{aligned}
        \widehat{R}_{pu}(g) = &\frac{\pi_p}{n_p} { \sum\limits_{i=1}^{n_p} } l(g(x_i^{p}),+1) + \\
        &(\frac{1}{n_u} { \sum\limits_{i=1}^{n_u} } l(g(x_i^{u}),-1) -
        \frac{\pi_p}{n_p} { \sum\limits_{i=1}^{n_p} } l(g(x_i^{p}),-1))
    \label{upu2}
    \end{aligned}
\end{equation}
where $n_p$ represents the number of labeled positive instances, $n_u$ represents the number of unlabeled instances, $x_i^p$ and $x_i^u$ denote instances in the labeled positive set and the unlabeled set, respectively. Theoretically, the second term in Eq.\ref{upu2} should be non-negative because that is used to estimate risk in negative part.
 Thus, \cite{kiryo2017positive} proposed non-negative PU (nnPU) methods by optimizing the following non-negative risk
 \begin{equation}
    \begin{aligned}
        \widetilde{R}_{pu}(g) = &\frac{\pi_p}{n_p} { \sum\limits_{i=1}^{n_p} } l(g(x_i^{p}),+1) + \\
        &max(0,\frac{1}{n_u} { \sum\limits_{i=1}^{n_u} } l(g(x_i^{u}),-1) -
        \frac{\pi_p}{n_p} { \sum\limits_{i=1}^{n_p} } l(g(x_i^{p}),-1))
    \label{nnpu}
    \end{aligned}
\end{equation}
From then on, many algorithms \cite{chen2020self, xu2019revisiting} for PU learning are proposed based upon nnPU.

\subsection{PUtree Framework}

\begin{figure*}[t]
  \centering
  \includegraphics[width=1.0\linewidth]{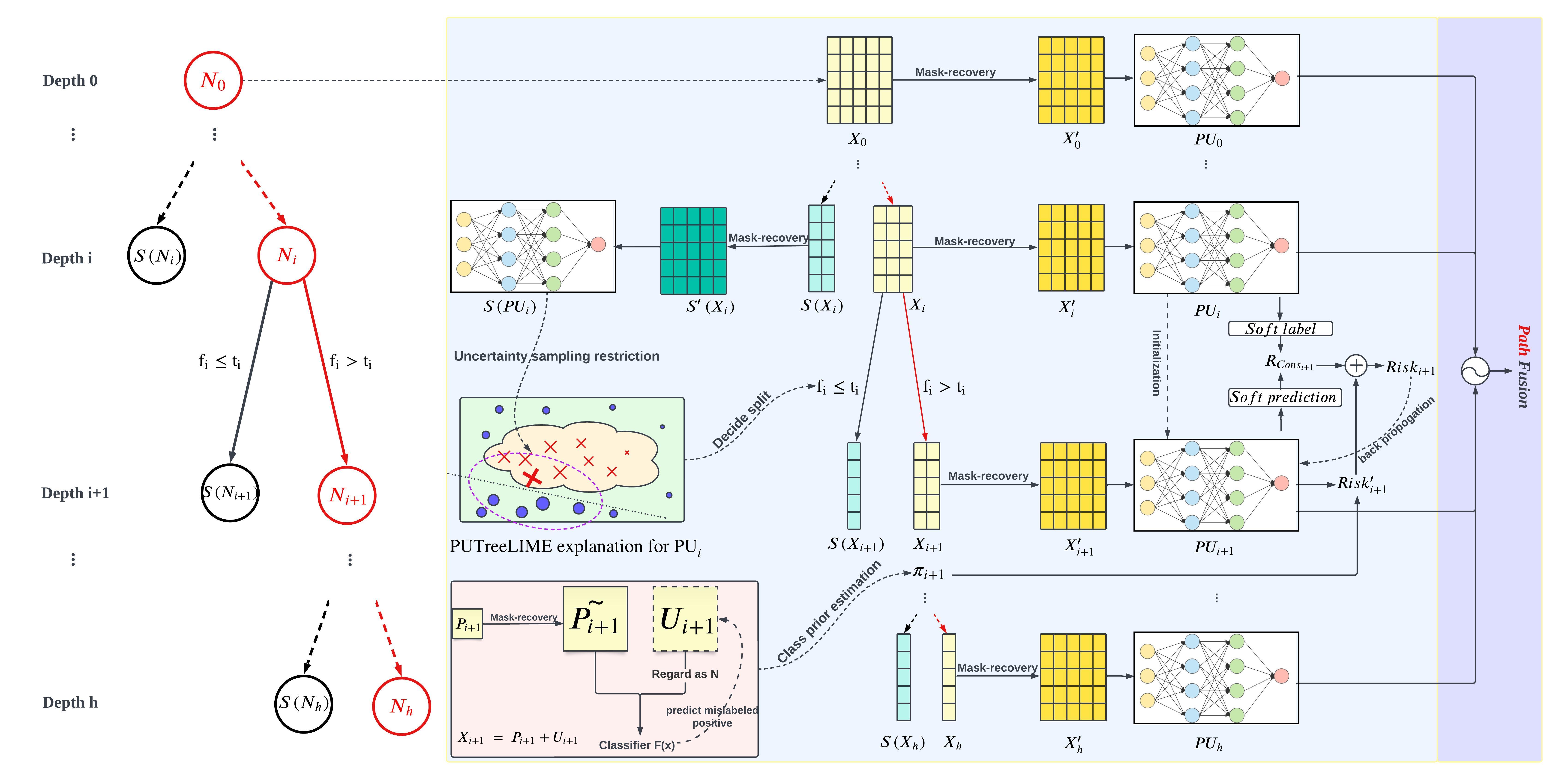}
  \caption{Illustration of our proposed PUtree network. We provide an example using the red tree path to showcase the PUtree algorithm. Specifically, we focus on the process of splitting from node $N_i$ to node $N_{i+1}$ and the corresponding sibling node represented by $S\left(N_{i+1}\right)$. Notably, when the red path extends to meet the termination criteria, hierarchical path fusion is conducted, seamlessly integrating information from the root to the leaf PU models. 
  \Description{Architecture of our proposed PU path fusion network.}}
  \label{model_overview}
\end{figure*}

In this section, we present an overview of PUtree framework, depicted in Figure \ref{model_overview}. The algorithm follows a top-to-down construction of the tree, where each node represents a community. We then perform hierarchical path fusion for each tree path. For illustration purposes, we use the red tree path in Figure \ref{model_overview} to demonstrate the process of tree splitting and the construction of nnPU models for each child-node (sub-community). Note that the PUtree structure is contingent upon the assigned task and training data utilized; meaning that the same dataset can require different tree structures when used to forecast various chronic illnesses. As is illustrated in Figure \ref{toy_example}, this is especially the case.

The construction of our PUtree begins with the root node $N_0$ at depth 0, where the training PU data $X_0$ is augmented using our mask-recovery data augmentation strategy to obtain augmented data $X_0^{'}$. Based on $X_0^{'}$, we build the corresponding nnPU model $PU_0$ using the nnPU algorithm \cite{kiryo2017positive}.

For each subsequent node $N_i$ (where $i$ represents the $i_{th}$ depth in the tree), we utilize the PUTreeLIME technique to explain the $PU_i$ model. This process allows us identify the optimal split criteria feature $f_i$ and its associated threshold $t_i$. The data $X_i$ is then split into two child-nodes: $N_{i+1}$ and $S(N_{i+1})$, where $S(N_{i+1})$ represents the sibling node of $N_{i+1}$. For both nodes, we follow the same procedure. Firstly, we apply mask-recovery data augmentation to enhance the learning effectiveness of $N_{i+1}$. Next, we utilize our enhanced class prior estimation method to predict the class prior $\pi_{i+1}$ (e.g., community disease rate), which remains consistent with $\pi_{i+1}^{'}$ due to the proportional augmentation of positive (P) and unlabeled (U) instances. Thereafter, we construct the nnPU model $PU_{i+1}$ using our proposed adversarial risk estimator, ensuring prediction consistency between $PU_{i+1}$ and its parent node model $PU_{i}$. This consistency fosters predictions that are not significantly divergent between sub-community models and their upper-level community models. The process is repeated until the termination criterion is met. Finally, hierarchical model fusion is performed on each path to generate the final prediction for an instance landing on that specific path. The detailed implementation is depicted in Algorithm \ref{alg:1}. Similar to \cite{wilton2022positive}, the termination conditions for PUtree encompass reaching the maximum tree depth $h$, the node size dropping below a specified threshold $S_{\text{min}}$, or the node consisting exclusively of unlabeled data (i.e., it is considered pure).

Our proposed PUtree framework presents a novel and comprehensive approach to integrating all nodes through leveraging interactions among nodes with their sibling and parent nodes, resulting in a substantial improvement in overall training performance. One of the key strengths of the PUtree lies in its remarkable interpretability, which provides valuable insights into the decision-making process of the model. Moreover, the PUtree effectively bridges the gap between traditional tree-based algorithms and deep neural networks, offering exciting opportunities for further research and application advancements in the field.

\begin{algorithm}
	\renewcommand{\algorithmicrequire}{\textbf{Input:}}
	\renewcommand{\algorithmicensure}{\textbf{Output:}}
	\caption{PUtree}
	\label{alg:1}
	\begin{algorithmic}[1]
		\REQUIRE positive-unlabeled data $D$, node $k$, nnPU model on current node $M$, feature $f$
        \ENSURE final classifier set T
        \STATE Initial T as empty list
        \IF{termination criterion is met}
            \STATE Train path fusion on current path get classifier $T^{'}$ 
            \STATE T$\leftarrow$ T + $T^{'}$\;
        \ELSE
            \STATE PUTreeLIME explains M to choose an optimal split (f,t)\;
            \STATE Create two child nodes $k_{f>t}$ and $k_{f\leq{t}}$for k\;

            \STATE Mask-recovery data augmentation for $D_{f>t}$ and $D_{f\leq{t}}$\;
            \STATE Positive class prior estimation for $D_{f>t}$ and $D_{f\leq{t}}$\;
            
            \STATE Build nnPU model for $D_{f>t}$ and $D_{f\leq{t}}$\;

            \STATE Learn PUtree($D_{f>t}$, $k_{f>t}$, $M_{f>t}$, f)
            \STATE Learn PUtree($D_{f\leq{t}}$, $f_{f\leq{t}}$, $M_{f\leq{t}}$, f)
        \ENDIF
		\STATE \textbf{Return} T
	\end{algorithmic}  
\end{algorithm}

\subsection{PUTreeLIME Split}


In order to better understand the proposed PUTreeLIME split method, we provide a brief introduction to Local Interpretable Model-Agnostic Explanations (LIME) \cite{ribeiro2016should}, which enables local explanations for complex models, including black-box algorithms. Formally, we denote the model to be explained as $f$, and the original instance to be locally explained as $x \in \mathbb{R}^{d}$. LIME uses $x' \in {0,1}^{d'}$ to represent an interpretable binary vector. For example, in the context of diabetes prediction, if the dimensions of $x$ represent features such as age and income, the dimensions of $x'$ would represent binary features such as \textit{age <= 60, age > 60, income <= 10k}, and \textit{income > 10k}. The thresholds for age and income can be obtained using a discretizer, such as quartiles or deciles, to convert the continuous data into binary features. Next, LIME randomly samples instances around $x'$ by drawing nonzero elements of $x'$ to obtain the set $z'$. The set $z'$ can then be recovered to $z \in \mathbb{R}^{d}$. Finally, the explainer model $g$ is trained using a weighted square loss.
 \begin{equation}
    \begin{aligned}
        L\left(f,g,d_x\right)\ =\ \sum\limits_{z,z'\in Z}d\left(x,z\right)\cdot\left(f\left(z\right)-g\left(z'\right)\right)^2
    \label{lime}
    \end{aligned}
\end{equation}


\noindent where $d(x,z)$ denotes the distance between the sampled instance $z$ and the original instance $x$, and $Z$ represents the set of sampled instances $z$ and $z'$.

Unlike LIME, we employ the uncertainty sampling method to sample instances $z$, which ensures that the sampled instances have a higher degree of responsibility for the model being explained. Specifically, when splitting node $N_i$ and explaining its model $PU_i$, we define the sampled set $Z$ as follows:

 \begin{equation}
    \begin{aligned}
        Z\ =Top_k\left(\underset{z\in D_{pool}}{\operatorname{argmax}} \quad a\left(z,\ S\left(PU_i\right)\right)\right)_{ }
    \label{sampling}
    \end{aligned}
\end{equation}

\noindent where $D_{pool}$ represents a large data pool of randomly sampled instances $z$ around the original instance, $S\left(PU_i\right)$ represents the nnPU model on the sibling node $S(N_{i})$, $k$ denotes the desired size of the set $Z$, and $a(.,.)$ represents an uncertainty sampling function, such as least confident sampling \cite{iyengar2000active}.

 \begin{equation}
    \begin{aligned}
        z_{LC}^-\ =\underset{z\in D_{pool}}{\operatorname{argmin}} \quad P_{\theta}\left(\overline{y} | z\right)
    \label{least}
    \end{aligned}
\end{equation}


\noindent where $\theta$ represents a well-trained model, $z_{LC}^-$ refers to the instance with the least confidence explained by $\theta$. Additionally, $\overline{y}$ represents the class with the largest predicted probability score for $z$.

 \begin{equation}
    \begin{aligned}
        \overline{y}\ =\underset{y\in C}{\operatorname{argmax}} \quad P_{\theta}\left(y | z \right)
    \label{least}
    \end{aligned}
\end{equation}

Eq. \ref{sampling} indicates that the sampled instance $z$ used to explain $PU_i$ is distinct from the instances in the node's sibling node, thereby enhancing the quality of the split for $PU_i$ and facilitating the selection of an optimal split criteria. This approach improves the effectiveness of the splitting process within the PUtree. The objective function for training PUTreeLIME is the same as that shown in Eq. \ref{lime}. After training the explainer model $g$, we define the optimal split criteria for the PUtree as follows:

 \begin{equation}
    \begin{aligned}
        \psi = 
        \underset{\beta\in\theta_g}{\operatorname{argmax}} \quad \left(KL\left(\tau^{i+1},S\left(\tau\right)^{i+1}\right) -\delta\left(B_{\left(N_{i+1}\right)}-B_{s\left(N_{i+1}\right)}\right)\right) \bigg| \beta
    \label{split}
    \end{aligned}
\end{equation}

\noindent where $\beta$ represents the feature coefficients in the explainer model $g$, and $\psi$ denotes the optimal split based on the corresponding feature with $\beta$. The term $\tau^{i+1}$ represents the predicted PN (positive-negative) distribution in node $N_{i+1}$, while $S\left(\tau\right)^{i+1}$ represents the predicted PN distribution in the sibling node $S(N_{i+1})$. Additionally, $B_{\left(N_{i+1}\right)}$ represents the number of instances in node $N_{i+1}$, and $B_{s\left(N_{i+1}\right)}$ represents the number of instances in the sibling node $S(N_{i+1})$. Finally, the term $\delta$ corresponds to the logistic function, which scales the second term in the equation to a range between 0 and 1. 

The KL divergence term in the optimal split equation serves to maximize the dissimilarity between the node $N_{i+1}$ and its sibling node, facilitating meaningful and informative splits within the PUtree framework. By emphasizing split fidelity, the algorithm can effectively capture distinctive patterns and separate instances based on their characteristics. Furthermore, the subtraction of the second term addresses the challenge of tree imbalance by promoting a balanced growth. This term helps control the difference in the number of instances between the newly split nodes, ensuring that the overall tree structure remains stable and avoids excessive disparities.

The optimal split equation, as represented by Eq. \ref{split}, strikes a delicate balance between split fidelity and maintaining a balanced tree structure. This balance enables the PUtree algorithm to achieve accurate and interpretable results, making it a powerful and robust approach for positive-unlabeled learning tasks. By effectively handling both the quality of splits and the growth of the tree, PUtree offers improved performance and enhanced interpretability, bridging the gap between traditional tree-based algorithms and deep neural networks.

\subsection{Child-Node PU Model Generation}
\subsubsection{Mask-recovery Data Augmentation}


To mitigate the potential underfitting concern that arises from the decreasing number of instances in deeper nodes of the tree, we propose a novel mask-recovery data augmentation strategy. Taking inspiration from the widely acclaimed BERT model \cite{devlin2018bert}, which employs masked language modeling to predict missing tokens, our approach applies a similar concept to feature masking.

Specifically, we identify the top $15\%$ most important features (for the target PU model) in each instance and mask them by replacing their values with placeholders. This masking procedure simulates the absence of critical information and challenges the model to infer or recover the masked features using the remaining ones. To accomplish this, we leverage a multi-label regression model \cite{xu2013multi} that takes the non-masked features as inputs and predicts the values of the masked features. By employing this mask-recovery process, we effectively generate augmented data that captures a broader range of feature interactions and potential patterns.

To enhance the diversity and quality of the augmented instances, we employ perturbations \cite{kargupta2003privacy} to the original data during the recovery process. These perturbations introduce subtle variations and noise, which further enrich the training samples and encourage the model to learn more robust and generalized representations. By expanding the training dataset with these high-quality augmented instances, our mask-recovery data augmentation strategy mitigates the risk of underfitting in the deeper nodes of the tree.

\subsubsection{PU Class Prior Estimation}

Estimating the positive class prior (e.g., diabetes rate for the target community) traditionally involves constructing a coarse binary classifier that assumes all unlabeled data is negative. This classifier assigns negative prediction scores to the original unlabeled instances. The subsequent step is to compare these scores with the mean score of predicted negative instances, known as $p_{\text{mean}}$. Instances with prediction scores higher than $p_{\text{mean}}$ are deemed reliable negatives, allowing us to estimate the proportion of positives in the unlabeled data. However, this approach suffers from a crucial drawback: it imposes an imbalance in the classification process by assuming all unlabeled instances are negative. This imbalance, exacerbated by the limited number of labeled positive examples in PU learning, can undermine the accuracy of the estimation.

The proposed mask-recovery data augmentation strategy can efficiently mitigate this issue, showcased in Figure \ref{model_overview}. Our data augmentation strategy empowers the classifier by augmenting the labeled positive data, thereby enhancing its stability and precision in estimating the positive class prior within the PU learning framework. Through this approach, we achieve a more refined and accurate understanding of the positive class distribution, enabling improved performance and reliability in PU learning tasks.
\subsubsection{Adversarial PU Risk Estimator}

PUtree algorithm divides the original PU data into distinct communities, leveraging the interactions among these communities to enhance the training process for each community model. To ensure the stability and capture the hierarchical relations within the PUtree, we propose an adversarial PU risk estimator for training each node's PU model. Inspired by knowledge distillation \cite{hinton2015distilling}, we employ the PU model from the parent node as a soft label provider, establishing consistency between the PU models at different levels. This formulation promotes knowledge transfer and enhances the overall performance of the PUtree algorithm.
\begin{equation}
    \begin{aligned}
        R_{Cons_{i}} = \frac{1}{N} \sum\limits_{{x\in X_i}} \left( PU_i(x) - PU_{i-1}(x)\right) ^2
    \label{Qi}
    \end{aligned}
\end{equation}
The variables $PU_i(x)$ and $PU_{i-1}(x)$ represent the prediction results for instance $x$ by $PU_i$ and $PU_{i-1}$, respectively. The variable $N$ denotes the number of instances in the data $X_i$ on the node. Building upon the nnPU risk estimator proposed by \cite{kiryo2017positive}, we introduce an adversarial PU risk estimator for $PU_i$, which is formulated as follows:
\begin{equation}
    \begin{aligned}
        Risk_{PU_i} = {Risk'}_{PU_i} + \lambda R_{Cons_{i}}
    \label{nnpu___ad}
    \end{aligned}
\end{equation}

where ${Risk'}_{PU_i}$ represents the same risk estimator as described in Eq. \ref{nnpu}, and $\lambda$ controls the contribution of consistency in the adversarial loss for $PU_i$.

The proposed adversarial PU risk estimator for $PU_i$ training leverages the concept of knowledge distillation and incorporates soft labels provided by the parent node model $PU_{i-1}$. This innovative approach enhances the consistency between the PU model on the current node and its parent node. By incorporating hierarchical relations into the training process, our method ensures improved stability and performance of the PUtree algorithm.

\subsection{Hierarchical Path Fusion Network}


Model fusion is a well-established strategy that combines information from multiple PU models (on PUtree) to make the prediction \cite{huang2020fusion}. This approach has found widespread applications in various domains, including electronic health record analysis \cite{huang2020fusion}, temperature forecasting \cite{lenartz2010enhanced}, and movie recommendation \cite{delen2007movie}. While fusion methods have been extensively studied in supervised learning scenarios, their application in the PU learning setting remains relatively unexplored.

To bridge this gap, we propose a novel PU fusion network designed specifically for the PUtree framework. Our approach leverages the hierarchical structure of the PUtree, capturing the relationships among communities at different depths. By aggregating the models on each path within the PUtree, our fusion network can effectively integrate information from multiple levels of the hierarchy. This enables the network to benefit from the collective knowledge and insights derived from each community, resulting in robust and accurate predictions.

When an instance $x$ lands on a path $\left\{PU_0,PU_1,...,PU_h\right\}$ on the PUtree, we feed it into the corresponding PU learning models $PU_i \in$ \\$\left\{PU_0,PU_1,...,PU_h\right\}$, where i is the depth of node. This allows us to obtain the last hidden layer representation $Q_i$ for the instance, which is computed as follows:
\begin{equation}
    \begin{aligned}
        Q_i =  PU_i(x)
    \label{Q1}
    \end{aligned}
\end{equation}

All the representations $Q_{i}$ are passed through a gated layer to obtain a comprehensive embedding that contains the aggregated information from the nnPU models at each depth. Inspired by the properties of GRU \cite{cho2014properties} and GMU \cite{arevalo2017gated}, we utilize a neuron $\phi$ with a ReLU function to encode the information from $PU_i$ for each representation $Q_{i}$. Additionally, a gate neuron $\sigma$ with a Sigmoid function is employed to control the contribution weight $K_i$ in the fusion process for each $Q_i$. The gate neuron specific to depth $i$ receives the concatenated feature vector from all depths as input and determines the significance of depth $i$ in the overall fusion. The equations for the fusion gate are as follows:
\begin{equation}
    \begin{aligned}
        V_i =  \phi(W_{V}^{i} \cdot Q_i)
    \label{Qi}
    \end{aligned}
\end{equation}
\begin{equation}
    \begin{aligned}
        K_i =  \sigma(W_{K}^{i} \cdot [Q_0,Q_1,...,Q_h])
    \label{Qi}
    \end{aligned}
\end{equation}
\begin{equation}
    \begin{aligned}
        Q^{H} =  \sum\limits_{i=0}^{h} K_{i} * V_{i}
    \label{Qi}
    \end{aligned}
\end{equation}
Furthermore, the fusion representation $Q^{H}$ is passed through a linear layer to obtain the fusion output $y^{H}$.
\begin{equation}
    \begin{aligned}
        y^{H} =  \phi(W_H \cdot  Q^{H})
    \label{Qi}
    \end{aligned}
\end{equation}
Following the approach of nnPU \cite{kiryo2017positive}, we utilize a non-negative risk estimator as our hierarchical path fusion risk estimator.
\begin{equation}
    \begin{aligned}
        R_H = &\frac{\pi_p}{n_p} { \sum\limits_{i=1}^{n_p} } l(F(x_i^{p}),+1) + \\
        &max(0,\frac{1}{n_u} { \sum\limits_{i=1}^{n_u} } l(F(x_i^{u}),-1) -
        \frac{\pi_p}{n_p} { \sum\limits_{i=1}^{n_p} } l(F(x_i^{p}),-1))
    \label{nnpu_fusion}
    \end{aligned}
\end{equation}
Here, $F$ represents the hierarchical path fusion model, and $R_H$ corresponds to the overall risk estimator on the path.

Overall, the proposed PU fusion network represents an innovative step towards exploring fusion strategies in the PU learning context. By capturing the hierarchical relations among communities and aggregating models on each path, our approach enhances the prediction performance of the PUtree, advancing the field of PU learning and paving the way for further research in this area.

\section{EXPERIMENT}
\subsection{Experimental settings}

\textbf{Datasets.} We introduced a new diabetes-screening dataset\footnote{https://github.com/YANGWU001/Putree/tree/main/dataset} that contains information on the shopping behavior of 53,764 individuals over a one-year period in China. The dataset includes attributes such as \textit{dairy purchase frequency}, \textit{meat purchase frequency}, etc. In this dataset, individuals were labeled as positive instances if they purchased diabetes-related products, including \textit{insulin}, \textit{syringes}, \textit{blood glucose test strips}, and \textit{diabetes urine test strips}. Out of the total instances, 4,496 are labeled as positive (8.4\%), while the remaining instances are used as unlabeled instance. To simulate the fact that approximately 12.4\% of whole population in China have been diagnosed with diabetes \cite{tu2022prevalence}, we set the positive class prior to 0.124 in both the train and test sets.
\begin{table*}[htp]\large
  \centering
  \caption{Summary of used datasets and their corresponding models.}
  \setlength{\tabcolsep}{1.3mm}{
    \begin{tabular}{ccccccccc}
    \hline
    Dataset & Dimension & $\# \text{Train}_{P}$ & $\# \text{Train}_{U}$  &   $ \text{Train}_{\pi_{P}}$ & \# Test&   $ \text{Test}_{\pi_{P}}$ & Positive Class & Model \\
    \hline
    Diabetes & 115 & 100 & 25,450 & 0.124 & 10,000 & 0.124  &  diabetes disease & 6-layer MLP \\
    UNSW-NB15 & 42 & 100   & 40,000 & 0.5 & 20,000 & 0.5  & attacks (e.g., DoS, worms) & 6-layer MLP \\
    NSL-KDD & 41 & 100 & 40,000 & 0.5 & 20,000 & 0.5  & attacks (e.g., DoS, R2L, U2R)& 6-layer MLP \\
    \hline
    
    \end{tabular}%
    }
  \label{tab:summary}%
\end{table*}%
\begin{table*}[htp]\large
  \centering
  \caption{Classification comparison on diabetes-prediction, UNSW-NB15 and NSL-KDD datasets. \best{Best} and \secondb{second} values are hightlighted as both highlighted. For each cell `x(y)', x, y denote the average value and standard deviation, respectively.}
  \setlength{\tabcolsep}{1.5mm}{
  \begin{tabular}{c|c|ccccc}
    \hline
    Dataset & Method  & Accuracy (\%) & Precision (\%) & Recall (\%) & F1 (\%) & F2 (\%)\\
    \hline
    \multicolumn{1}{c|}{\multirow{4}{*}{Diabetes}} & naive & \best{87.76 (0.03)} & \best{74.60 (5.55)} & 1.92 (0.17) & 3.74 (0.32)& 2.38 (0.21)\\
          & uPU   & 83.85 (0.18) & 29.58 (0.77) & 21.93 (1.79) & 25.17 (1.42) & 23.12 (1.67)\\
          & nnPU  & 84.20 (0.69) & 34.38 (0.99) & \secondb{30.02 (4.32)} & 31.88 (2.49)& \secondb{30.71 (3.65)}\\
          & VPU  & \secondb{85.98 (0.78)} & \secondb{42.05 (2.52)} & 27.41 (2.17) & \secondb{33.12 (1.76)}& 29.43 (2.02)\\
    \cline{2-7} & PUtree & 81.07 (1.57) & 31.77 (1.56) & \best{45.18 (5.52)} & \best{37.11 (1.28)}& \best{41.66 (3.66)}\\
    \hline
    \multicolumn{1}{c|}{\multirow{4}{*}{UNSW-NB15}} & naive & 73.45 (1.61) & 79.86 (2.71) & 62.84 (3.72) & 70.26 (2.29)& 65.60 (3.16)\\
          & uPU  & 76.11 (1.05) & \best{87.49 (2.37)} & 61.01 (2.38) & 71.85 (1.52) & 64.92 (2.08)\\
          & nnPU  & 80.68 (0.78) & 81.37 (1.51) & \secondb{79.64 (1.94)} & 80.47 (0.83)& \secondb{79.91 (1.41)}\\
          & VPU  & \secondb{81.20 (0.47)} & \secondb{83.77 (6.59)} & 79.15 (11.26) & \secondb{80.61 (2.27)}& 79.54 (7.57)\\
          \cline{2-7} & PUtree  & \best{83.49 (0.50)} & 79.60 (1.11) & \best{90.09 (1.55)} & \best{84.51 (0.44)}& \best{87.77 (0.98)}\\
    \hline

        \multicolumn{1}{c|}{\multirow{4}{*}{NSL-KDD}} & naive & 77.34 (0.69) & 87.35 (1.57) & 63.97 (2.27) & 73.82 (1.19)& 67.57 (1.91)\\
          & uPU   & 84.14 (1.34) & \best{97.31 (0.77)} & 70.23 (2.98) & 81.55 (1.92) & 74.35 (2.64)\\
          & nnPU  & 89.62 (0.80) & 93.45 (0.58) & \secondb{85.21 (1.57)} & 89.14 (0.91)& \secondb{86.74 (1.31)}\\
          & VPU  & \secondb{90.30 (0.64)} & \secondb{95.93 (1.31)} & 84.04 (1.33) & \secondb{89.58 (0.69)}& 86.17 (1.03)\\
    \cline{2-7} & PUtree & \best{91.34 (0.83)} & 89.91 (1.66) & \best{92.42 (2.13)} & \best{91.12 (0.85)}& \best{91.89 (1.49)}\\
    \hline
    \end{tabular}
    }
  \label{tab:com_results}
\end{table*}
Additionally, to evaluate the generalizability of the proposed PUtree model, we conducted experiments on two PU-learning benchmarks: UNSW-NB15 \cite{moustafa2015unsw} and NSL-KDD \cite{dhanabal2015study}. These benchmarks are widely used in intrusion detection tasks for network security. They consist of instances representing normal network behavior as well as various types of attacks. The positive class represents the attacks, while the negative class represents normal instances. For each benchmark, following other PU baselines, we randomly selected 20,000 positive instances and 20,000 negative instances to form the unlabeled set. In the training data, we included 100 labeled positive instances, following the setup used by \cite{kiryo2017positive}. For the test set, 10,000 positive and 10,000 negative instances were randomly sampled from each benchmark. Table \ref{tab:summary} provides an overview of the statistical details of the datasets, including the number of instances, positive instances, negative instances, and other relevant information.

\textbf{Implementation details.} \footnote{https://github.com/YANGWU001/Putree} Following the approach of nnPU \cite{kiryo2017positive}, a 6-layer multilayer perceptron (MLP) with ReLU activation was employed in the nnPU model. During the training process, the Adap optimizer was utilized with a learning rate of $1e^{-4}$, and a batch size of 4000 was employed for all datasets. All experiments were conducted using a NVIDIA V100 GPU, with training conducted for 100 epochs.

\textbf{Evaluation metrics.} The performance of the model was assessed across five metrics on the test set, including Accuracy, Precision, Recall, F1 score, and F2 score. Given that the misclassification of positive cases as negative can lead to severe repercussions, such as postponed medical care \cite{wang2022adaptive}, special emphasis was placed on the F2 score for diabetes prediction. The F2 score is exclusive in its consideration of True Positive determinations. Additionally, to ensure a fair comparison, each experiment is conducted five times, and the mean and standard deviations of each metric are reported. This approach helps to account for any variations or fluctuations in the model's performance across multiple runs, providing a more comprehensive assessment of its effectiveness and stableness.\\

\subsection{Overall Comparisons}
\textbf{Baselines.} We compare PUtree with the following baseline models:\vspace{2pt}\\
$\bullet$ \textbf{Naive}. The naive method treats all unlabeled instance as negative data and then builds a decision tree model.\\
$\bullet$ \textbf{uPU} \cite{du2015convex}. Unbiased Positive-unlabeled learning, which is a classic cost-sensitive learning approach learning positive and unlabel data based on unbiased risk estimators.\\
$\bullet$ \textbf{nnPU} \cite{kiryo2017positive}. None-Negative Positive-unlabeled learning addresses the overfitting issue in uPU by ensuring non-negativity of the empirical risk on the negative part. \\
$\bullet$ \textbf{VPU} \cite{chen2020variational}. Variational PU (VPU) is a novel PU learning method that utilizes a variational principle to evaluate modeling error directly from given data without the need for class prior estimation.

\begin{table*}[htp]\large
  \centering
  \caption{Ablation results on UNSW-NB15 with \cmark indicating the enabling of the corresponding module.}
   \setlength{\tabcolsep}{1.5mm}{
    \begin{tabular}{c|cccc|ccccc}
    \hline
    Variant & PUTL & MR & PF & AR & Accuracy (\%) & Precision (\%) & Recall (\%) & F1 (\%)& F2 (\%)\\
    \hline
    I     & \xmark     & \cmark & \cmark & \cmark  & 81.79 (0.39) & 78.88 (2.74) & 87.23 (5.32) & 82.69 (1.01) & 85.31 (3.48)    \\
    II    & \cmark     & \xmark    & \cmark & \cmark  & 81.19 (1.16) & \best{81.40 (5.14)} & 81.88 (8.87) & 81.15 (2.71) & 81.49 (6.52) \\
    III   & \cmark & \cmark  &\xmark     & \cmark    & \secondb{82.41 (0.54)} & 77.05 (0.81) & \best{92.33 (1.39)} & \secondb{83.99 (0.51)}   & \best{88.80 (0.94)} \\
    IV   & \cmark & \cmark  &\cmark     & \xmark   & 81.07 (0.27) & 78.81 (1.34) & 85.10 (2.91) & 81.79 (0.66)  & 83.73 (1.96)   \\
    \hline
    PUtree    & \cmark     & \cmark  & \cmark& \cmark   & \best{83.49 (0.50)} & \secondb{79.60 (1.11)} & \secondb{90.09 (1.55)} & \best{84.51 (0.44)} & \secondb{87.77 (0.98)}  \\
    \hline
    \end{tabular}
    }
  \label{tab:ablation}%
\end{table*}

\textbf{Results.} Table \ref{tab:com_results} provides a comprehensive summary of the overall performance comparison results. The proposed PUtree model demonstrates significant superiority over competitors across all datasets (\textit{p < 0.05}), exhibiting remarkable improvements in most metrics. Specifically, for the diabetes-prediction dataset, PUtree surpasses the second-best competitor by approximately 15\% in recall, 4\% in F1 score, and 11\% in F2 score. Moreover, PUtree achieves notable enhancements of 1\% to 2\% in terms of accuracy for the UNSW-NB15 and NSL-KDD datasets. This validates the effectiveness of our proposed method. Our PUtree algorithm exhibits remarkable superiority compared to other algorithms for diabetes prediction, delivering exceptional performance in terms of recall, F1 score, and F2 score. This outcome aligns precisely with our expectations in identifying individuals at a high risk of diabetes. Moreover, the precision achieved by PUtree remains consistently higher than the extremely low prior probability of randomly selecting a positive instance from the PU data, which accounts for the 12.4\% prevalence of diabetes. This finding underscores the ability of PUtree to provide a stable and dependable prediction for diabetes diagnosis. Furthermore, VPU is considered as one of the most competitive baselines. It leverages a variational principle to directly evaluate modeling error from the given data, enabling efficient optimization of binary classifiers trained on positive and unlabeled data. Nevertheless, our PUtree model still exhibits more robust performance compared to the VPU method.\\

\subsection{Ablation Study}

To gain a comprehensive investigation of the impact of each module in PUtree, namely PUTreeLIME (PUTL), mask-recovery data augmentation (MR), path fusion (PF), and adversarial risk estimator (AR), we conducted ablation studies on the UNSW-NB15 dataset. By comparing PUtree with its four variants, we can explore the importance of each module:\\
$\bullet$ Variant I: In this variant, when splitting nodes, PUTreeLIME is not used, and instead, LIME is employed for node splitting.\\
$\bullet$ Variant II: This variant does not utilize the mask-recovery data augmentation strategy. Instead, the original data is used to build the PU model for each node.\\
$\bullet$ Variant III: Instead of using the fusioned path model to make the final prediction, this variant directly utilizes the leaf node to make predictions.\\
$\bullet$ Variant 4: In this variant, when building child node PU models, the effect of the parent node is not taken into consideration. Instead, the child model is trained using the nnPU \cite{kiryo2017positive} risk estimator directly.\\
Comparing PUtree with its variants reveals the crucial role each module plays in our PUtree framework. The results are shown in Table \ref{tab:ablation}. PUtree exhibits robustness across all variants, achieving an accuracy of 83.49\% and an F1 score of 84.51\%. Notably, in the comparison of variant IV with PUtree, we observe the significant impact of the adversarial loss in capturing the hierarchical relation within PUtree, leading to a substantial enhancement in overall performance. The comparison between variant I and PUtree demonstrates the effectiveness of PUtreeLIME in the PU split setting. The comparison between variant II and PUtree emphasizes the significant improvement achieved by the mask-recovery data augmentation strategy in PUtree. Lastly, the comparison between variant III and PUtree indicates that path fusion outperforms single leaf node predictions, showcasing the robustness of this approach.

\begin{figure}[t]
  \centering
  \includegraphics[width=\linewidth]{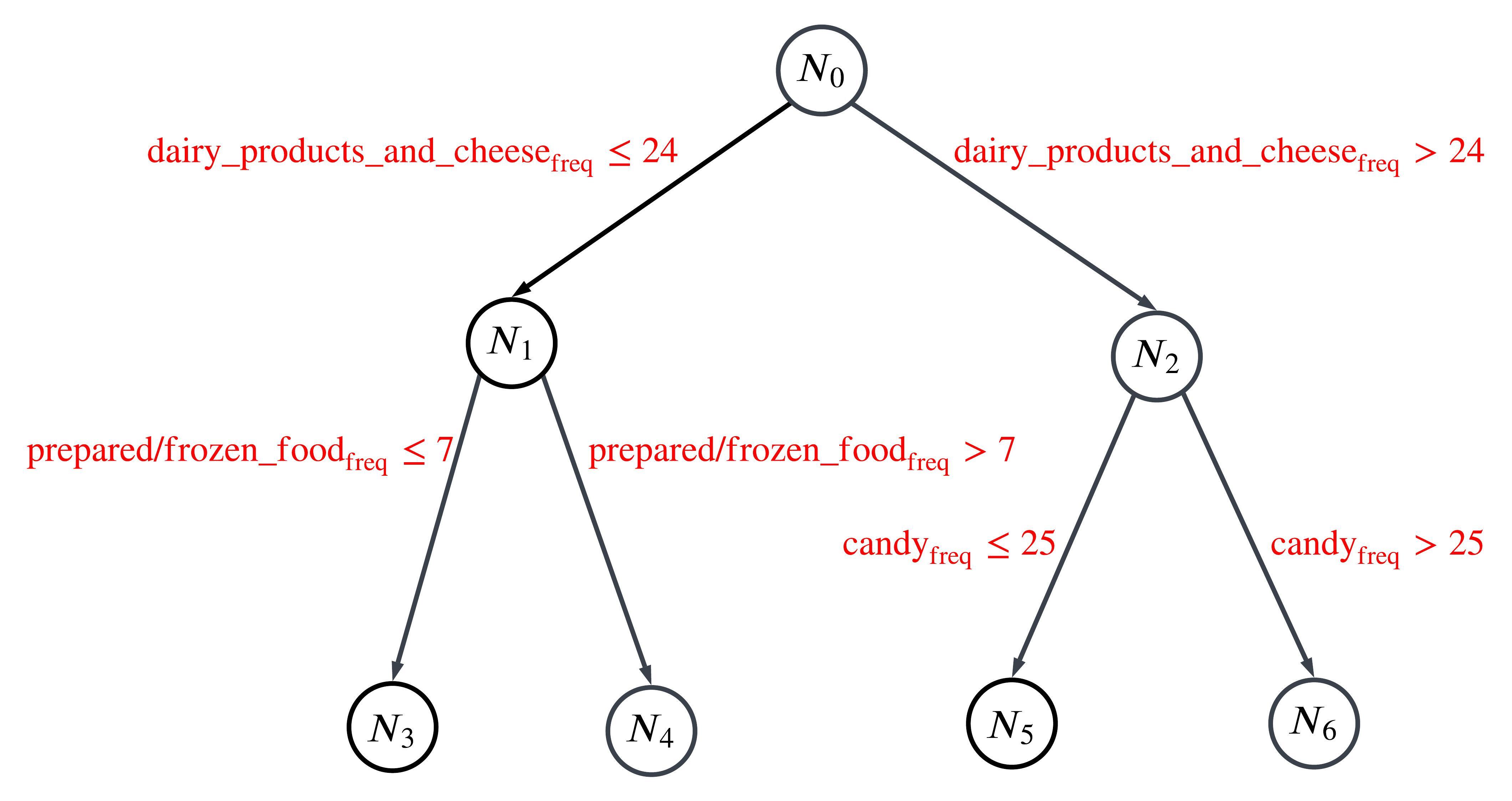}
  \caption{A PUtree example for diabetes-prediction dataset. Where $N_0$ represents the whole population, and the population is consistently split into sub-communities based on our proposed PUTreeLIME. `$\rm{product}_{freq}$' represents the annual purchase frequency of the product.}
  \Description{The framework of our proposed PU path fusion network.}
  \label{putree_example}
\end{figure}
\subsection{PUtree Analysis}
The interpretability of PUtree is demonstrated in Figure \ref{putree_example}, which contains a small sub-PUtree generated from the diabetes-prediction dataset. The findings of this tree structure reveal that an increased purchase frequency of cheese, consumption of prepared/frozen foods, and a high intake of candy are correlated with an elevated risk of diabetes. These results corroborate the recent evidence suggesting that dairy, processed products, and sugars can alter the glucose metabolism patterns and potentially result in type 2 diabetes. \cite{alvarez2019effects, gao2013dairy}.



\section{CONCLUSION AND FUTURE WORK}

In conclusion, this work has demonstrated the potential of applying multi-PU model fusion within the context of chronic disease prediction/screening, enabling us to analyze differences among diverse populations in a way that was previously unfeasible. The PUtree model has provided a generalizable method for efficiently characterizing PU models in any algorithm-detected population, as well as considering hierarchical relationships among them. The proposed PUTreeLIME method enhances interpretability and usability by considering the interaction with sibling communities. Additionally, by utilizing hierarchical PU model fusion on the PUtree path, we are able to make more accurate forecasts which leverage a variety of zoom-in demographics or behavior characteristics of the general population. Overall, the proposed model advances the field of PU learning to address the challenge of chronic disease detection, which, as experiment shows, can be generalized to other PU learning scenarios. In future, we will further optimize the data augmentation method, and conduct more experiments to validate and explore its effectiveness in diverse real-world scenarios.

\bibliographystyle{ACM-Reference-Format}
\bibliography{sample-base}


\begin{thebibliography}{49}


\ifx \showCODEN    \undefined \def \showCODEN     #1{\unskip}     \fi
\ifx \showDOI      \undefined \def \showDOI       #1{#1}\fi
\ifx \showISBNx    \undefined \def \showISBNx     #1{\unskip}     \fi
\ifx \showISBNxiii \undefined \def \showISBNxiii  #1{\unskip}     \fi
\ifx \showISSN     \undefined \def \showISSN      #1{\unskip}     \fi
\ifx \showLCCN     \undefined \def \showLCCN      #1{\unskip}     \fi
\ifx \shownote     \undefined \def \shownote      #1{#1}          \fi
\ifx \showarticletitle \undefined \def \showarticletitle #1{#1}   \fi
\ifx \showURL      \undefined \def \showURL       {\relax}        \fi
\providecommand\bibfield[2]{#2}
\providecommand\bibinfo[2]{#2}
\providecommand\natexlab[1]{#1}
\providecommand\showeprint[2][]{arXiv:#2}

\bibitem[Alvarez-Bueno et~al\mbox{.}(2019)]%
        {alvarez2019effects}
\bibfield{author}{\bibinfo{person}{Celia Alvarez-Bueno}, \bibinfo{person}{Ivan
  Cavero-Redondo}, \bibinfo{person}{Vicente Martinez-Vizcaino},
  \bibinfo{person}{Mercedes Sotos-Prieto}, \bibinfo{person}{Jonatan~R Ruiz},
  {and} \bibinfo{person}{Angel Gil}.} \bibinfo{year}{2019}\natexlab{}.
\newblock \showarticletitle{Effects of milk and dairy product consumption on
  type 2 diabetes: overview of systematic reviews and meta-analyses}.
\newblock \bibinfo{journal}{\emph{Advances in Nutrition}} \bibinfo{volume}{10},
  \bibinfo{number}{suppl\_2} (\bibinfo{year}{2019}),
  \bibinfo{pages}{S154--S163}.
\newblock


\bibitem[Anaby-Tavor et~al\mbox{.}(2020)]%
        {anaby2020not}
\bibfield{author}{\bibinfo{person}{Ateret Anaby-Tavor}, \bibinfo{person}{Boaz
  Carmeli}, \bibinfo{person}{Esther Goldbraich}, \bibinfo{person}{Amir Kantor},
  \bibinfo{person}{George Kour}, \bibinfo{person}{Segev Shlomov},
  \bibinfo{person}{Naama Tepper}, {and} \bibinfo{person}{Naama Zwerdling}.}
  \bibinfo{year}{2020}\natexlab{}.
\newblock \showarticletitle{Do not have enough data? Deep learning to the
  rescue!}. In \bibinfo{booktitle}{\emph{Proceedings of the AAAI Conference on
  Artificial Intelligence}}, Vol.~\bibinfo{volume}{34}.
  \bibinfo{pages}{7383--7390}.
\newblock


\bibitem[Arevalo et~al\mbox{.}(2017)]%
        {arevalo2017gated}
\bibfield{author}{\bibinfo{person}{John Arevalo}, \bibinfo{person}{Thamar
  Solorio}, \bibinfo{person}{Manuel Montes-y G{\'o}mez}, {and}
  \bibinfo{person}{Fabio~A Gonz{\'a}lez}.} \bibinfo{year}{2017}\natexlab{}.
\newblock \showarticletitle{Gated multimodal units for information fusion}.
\newblock \bibinfo{journal}{\emph{arXiv preprint arXiv:1702.01992}}
  (\bibinfo{year}{2017}).
\newblock


\bibitem[Bekker and Davis(2018)]%
        {bekker2018estimating}
\bibfield{author}{\bibinfo{person}{Jessa Bekker} {and} \bibinfo{person}{Jesse
  Davis}.} \bibinfo{year}{2018}\natexlab{}.
\newblock \showarticletitle{Estimating the class prior in positive and
  unlabeled data through decision tree induction}. In
  \bibinfo{booktitle}{\emph{Proceedings of the AAAI Conference on Artificial
  Intelligence}}, Vol.~\bibinfo{volume}{32}.
\newblock


\bibitem[Berkowitz et~al\mbox{.}(2013)]%
        {berkowitz2013clinical}
\bibfield{author}{\bibinfo{person}{Seth~A Berkowitz},
  \bibinfo{person}{Katherine Aragon}, \bibinfo{person}{Jonas Hines},
  \bibinfo{person}{Hilary Seligman}, \bibinfo{person}{Sei Lee}, {and}
  \bibinfo{person}{Urmimala Sarkar}.} \bibinfo{year}{2013}\natexlab{}.
\newblock \showarticletitle{Do clinical standards for diabetes care address
  excess risk for hypoglycemia in vulnerable patients? A systematic review}.
\newblock \bibinfo{journal}{\emph{Health services research}}
  \bibinfo{volume}{48}, \bibinfo{number}{4} (\bibinfo{year}{2013}),
  \bibinfo{pages}{1299--1310}.
\newblock


\bibitem[Breiman(2001)]%
        {breiman2001random}
\bibfield{author}{\bibinfo{person}{Leo Breiman}.}
  \bibinfo{year}{2001}\natexlab{}.
\newblock \showarticletitle{Random forests}.
\newblock \bibinfo{journal}{\emph{Machine learning}} \bibinfo{volume}{45},
  \bibinfo{number}{1} (\bibinfo{year}{2001}), \bibinfo{pages}{5--32}.
\newblock


\bibitem[Chang et~al\mbox{.}(2021)]%
        {chang2021positive}
\bibfield{author}{\bibinfo{person}{Shizhen Chang}, \bibinfo{person}{Bo Du},
  {and} \bibinfo{person}{Liangpei Zhang}.} \bibinfo{year}{2021}\natexlab{}.
\newblock \showarticletitle{Positive unlabeled learning with class-prior
  approximation}. In \bibinfo{booktitle}{\emph{Proceedings of the Twenty-Ninth
  International Conference on International Joint Conferences on Artificial
  Intelligence}}. \bibinfo{pages}{2014--2021}.
\newblock


\bibitem[Chen et~al\mbox{.}(2020b)]%
        {chen2020variational}
\bibfield{author}{\bibinfo{person}{Hui Chen}, \bibinfo{person}{Fangqing Liu},
  \bibinfo{person}{Yin Wang}, \bibinfo{person}{Liyue Zhao}, {and}
  \bibinfo{person}{Hao Wu}.} \bibinfo{year}{2020}\natexlab{b}.
\newblock \showarticletitle{A variational approach for learning from positive
  and unlabeled data}.
\newblock \bibinfo{journal}{\emph{Advances in Neural Information Processing
  Systems}}  \bibinfo{volume}{33} (\bibinfo{year}{2020}),
  \bibinfo{pages}{14844--14854}.
\newblock


\bibitem[Chen et~al\mbox{.}(2020a)]%
        {chen2020self}
\bibfield{author}{\bibinfo{person}{Xuxi Chen}, \bibinfo{person}{Wuyang Chen},
  \bibinfo{person}{Tianlong Chen}, \bibinfo{person}{Ye Yuan},
  \bibinfo{person}{Chen Gong}, \bibinfo{person}{Kewei Chen}, {and}
  \bibinfo{person}{Zhangyang Wang}.} \bibinfo{year}{2020}\natexlab{a}.
\newblock \showarticletitle{Self-pu: Self boosted and calibrated
  positive-unlabeled training}. In \bibinfo{booktitle}{\emph{International
  Conference on Machine Learning}}. PMLR, \bibinfo{pages}{1510--1519}.
\newblock


\bibitem[Cho et~al\mbox{.}(2014)]%
        {cho2014properties}
\bibfield{author}{\bibinfo{person}{Kyunghyun Cho}, \bibinfo{person}{Bart
  Van~Merri{\"e}nboer}, \bibinfo{person}{Dzmitry Bahdanau}, {and}
  \bibinfo{person}{Yoshua Bengio}.} \bibinfo{year}{2014}\natexlab{}.
\newblock \showarticletitle{On the properties of neural machine translation:
  Encoder-decoder approaches}.
\newblock \bibinfo{journal}{\emph{arXiv preprint arXiv:1409.1259}}
  (\bibinfo{year}{2014}).
\newblock


\bibitem[Delen et~al\mbox{.}(2007)]%
        {delen2007movie}
\bibfield{author}{\bibinfo{person}{Dursun Delen}, \bibinfo{person}{Ramesh
  Sharda}, {and} \bibinfo{person}{Prajeeb Kumar}.}
  \bibinfo{year}{2007}\natexlab{}.
\newblock \showarticletitle{Movie forecast Guru: A Web-based DSS for Hollywood
  managers}.
\newblock \bibinfo{journal}{\emph{Decision Support Systems}}
  \bibinfo{volume}{43}, \bibinfo{number}{4} (\bibinfo{year}{2007}),
  \bibinfo{pages}{1151--1170}.
\newblock


\bibitem[Denis(1998)]%
        {denis1998pac}
\bibfield{author}{\bibinfo{person}{Fran{\c{c}}ois Denis}.}
  \bibinfo{year}{1998}\natexlab{}.
\newblock \showarticletitle{PAC learning from positive statistical queries}. In
  \bibinfo{booktitle}{\emph{Algorithmic Learning Theory: 9th International
  Conference, ALT’98 Otzenhausen, Germany, October 8--10, 1998 Proceedings
  9}}. Springer, \bibinfo{pages}{112--126}.
\newblock


\bibitem[Devlin et~al\mbox{.}(2018)]%
        {devlin2018bert}
\bibfield{author}{\bibinfo{person}{Jacob Devlin}, \bibinfo{person}{Ming-Wei
  Chang}, \bibinfo{person}{Kenton Lee}, {and} \bibinfo{person}{Kristina
  Toutanova}.} \bibinfo{year}{2018}\natexlab{}.
\newblock \showarticletitle{Bert: Pre-training of deep bidirectional
  transformers for language understanding}.
\newblock \bibinfo{journal}{\emph{arXiv preprint arXiv:1810.04805}}
  (\bibinfo{year}{2018}).
\newblock


\bibitem[Dhanabal and Shantharajah(2015)]%
        {dhanabal2015study}
\bibfield{author}{\bibinfo{person}{L Dhanabal} {and} \bibinfo{person}{SP
  Shantharajah}.} \bibinfo{year}{2015}\natexlab{}.
\newblock \showarticletitle{A study on NSL-KDD dataset for intrusion detection
  system based on classification algorithms}.
\newblock \bibinfo{journal}{\emph{International journal of advanced research in
  computer and communication engineering}} \bibinfo{volume}{4},
  \bibinfo{number}{6} (\bibinfo{year}{2015}), \bibinfo{pages}{446--452}.
\newblock


\bibitem[Du~Plessis et~al\mbox{.}(2015)]%
        {du2015convex}
\bibfield{author}{\bibinfo{person}{Marthinus Du~Plessis}, \bibinfo{person}{Gang
  Niu}, {and} \bibinfo{person}{Masashi Sugiyama}.}
  \bibinfo{year}{2015}\natexlab{}.
\newblock \showarticletitle{Convex formulation for learning from positive and
  unlabeled data}. In \bibinfo{booktitle}{\emph{International conference on
  machine learning}}. PMLR, \bibinfo{pages}{1386--1394}.
\newblock


\bibitem[Du~Plessis et~al\mbox{.}(2014)]%
        {du2014analysis}
\bibfield{author}{\bibinfo{person}{Marthinus~C Du~Plessis},
  \bibinfo{person}{Gang Niu}, {and} \bibinfo{person}{Masashi Sugiyama}.}
  \bibinfo{year}{2014}\natexlab{}.
\newblock \showarticletitle{Analysis of learning from positive and unlabeled
  data}.
\newblock \bibinfo{journal}{\emph{Advances in neural information processing
  systems}}  \bibinfo{volume}{27} (\bibinfo{year}{2014}).
\newblock


\bibitem[Gao et~al\mbox{.}(2013)]%
        {gao2013dairy}
\bibfield{author}{\bibinfo{person}{Dengfeng Gao}, \bibinfo{person}{Ning Ning},
  \bibinfo{person}{Congxia Wang}, \bibinfo{person}{Yuhuan Wang},
  \bibinfo{person}{Qing Li}, \bibinfo{person}{Zhe Meng}, \bibinfo{person}{Yang
  Liu}, {and} \bibinfo{person}{Qiang Li}.} \bibinfo{year}{2013}\natexlab{}.
\newblock \showarticletitle{Dairy products consumption and risk of type 2
  diabetes: systematic review and dose-response meta-analysis}.
\newblock \bibinfo{journal}{\emph{PloS one}} \bibinfo{volume}{8},
  \bibinfo{number}{9} (\bibinfo{year}{2013}), \bibinfo{pages}{e73965}.
\newblock


\bibitem[Hinton et~al\mbox{.}(2015)]%
        {hinton2015distilling}
\bibfield{author}{\bibinfo{person}{Geoffrey Hinton}, \bibinfo{person}{Oriol
  Vinyals}, {and} \bibinfo{person}{Jeff Dean}.}
  \bibinfo{year}{2015}\natexlab{}.
\newblock \showarticletitle{Distilling the knowledge in a neural network}.
\newblock \bibinfo{journal}{\emph{arXiv preprint arXiv:1503.02531}}
  (\bibinfo{year}{2015}).
\newblock


\bibitem[Hou et~al\mbox{.}(2017)]%
        {hou2017generative}
\bibfield{author}{\bibinfo{person}{Ming Hou}, \bibinfo{person}{Brahim
  Chaib-Draa}, \bibinfo{person}{Chao Li}, {and} \bibinfo{person}{Qibin Zhao}.}
  \bibinfo{year}{2017}\natexlab{}.
\newblock \showarticletitle{Generative adversarial positive-unlabelled
  learning}.
\newblock \bibinfo{journal}{\emph{arXiv preprint arXiv:1711.08054}}
  (\bibinfo{year}{2017}).
\newblock


\bibitem[Hu et~al\mbox{.}(2021)]%
        {hu2021predictive}
\bibfield{author}{\bibinfo{person}{Wenpeng Hu}, \bibinfo{person}{Ran Le},
  \bibinfo{person}{Bing Liu}, \bibinfo{person}{Feng Ji},
  \bibinfo{person}{Jinwen Ma}, \bibinfo{person}{Dongyan Zhao}, {and}
  \bibinfo{person}{Rui Yan}.} \bibinfo{year}{2021}\natexlab{}.
\newblock \showarticletitle{Predictive adversarial learning from positive and
  unlabeled data}. In \bibinfo{booktitle}{\emph{Proceedings of the AAAI
  Conference on Artificial Intelligence}}, Vol.~\bibinfo{volume}{35}.
  \bibinfo{pages}{7806--7814}.
\newblock


\bibitem[Huang et~al\mbox{.}(2020)]%
        {huang2020fusion}
\bibfield{author}{\bibinfo{person}{Shih-Cheng Huang}, \bibinfo{person}{Anuj
  Pareek}, \bibinfo{person}{Saeed Seyyedi}, \bibinfo{person}{Imon Banerjee},
  {and} \bibinfo{person}{Matthew~P Lungren}.} \bibinfo{year}{2020}\natexlab{}.
\newblock \showarticletitle{Fusion of medical imaging and electronic health
  records using deep learning: a systematic review and implementation
  guidelines}.
\newblock \bibinfo{journal}{\emph{NPJ digital medicine}} \bibinfo{volume}{3},
  \bibinfo{number}{1} (\bibinfo{year}{2020}), \bibinfo{pages}{136}.
\newblock


\bibitem[Inoue(2018)]%
        {inoue2018data}
\bibfield{author}{\bibinfo{person}{Hiroshi Inoue}.}
  \bibinfo{year}{2018}\natexlab{}.
\newblock \showarticletitle{Data augmentation by pairing samples for images
  classification}.
\newblock \bibinfo{journal}{\emph{arXiv preprint arXiv:1801.02929}}
  (\bibinfo{year}{2018}).
\newblock


\bibitem[Iyengar et~al\mbox{.}(2000)]%
        {iyengar2000active}
\bibfield{author}{\bibinfo{person}{Vijay~S Iyengar}, \bibinfo{person}{Chidanand
  Apte}, {and} \bibinfo{person}{Tong Zhang}.} \bibinfo{year}{2000}\natexlab{}.
\newblock \showarticletitle{Active learning using adaptive resampling}. In
  \bibinfo{booktitle}{\emph{Proceedings of the sixth ACM SIGKDD international
  conference on Knowledge discovery and data mining}}. \bibinfo{pages}{91--98}.
\newblock


\bibitem[Kargupta et~al\mbox{.}(2003)]%
        {kargupta2003privacy}
\bibfield{author}{\bibinfo{person}{Hillol Kargupta}, \bibinfo{person}{Souptik
  Datta}, \bibinfo{person}{Qi Wang}, {and} \bibinfo{person}{Krishnamoorthy
  Sivakumar}.} \bibinfo{year}{2003}\natexlab{}.
\newblock \showarticletitle{On the privacy preserving properties of random data
  perturbation techniques}. In \bibinfo{booktitle}{\emph{Third IEEE
  international conference on data mining}}. IEEE, \bibinfo{pages}{99--106}.
\newblock


\bibitem[Kiryo et~al\mbox{.}(2017)]%
        {kiryo2017positive}
\bibfield{author}{\bibinfo{person}{Ryuichi Kiryo}, \bibinfo{person}{Gang Niu},
  \bibinfo{person}{Marthinus~C Du~Plessis}, {and} \bibinfo{person}{Masashi
  Sugiyama}.} \bibinfo{year}{2017}\natexlab{}.
\newblock \showarticletitle{Positive-unlabeled learning with non-negative risk
  estimator}.
\newblock \bibinfo{journal}{\emph{Advances in neural information processing
  systems}}  \bibinfo{volume}{30} (\bibinfo{year}{2017}).
\newblock


\bibitem[Lenartz et~al\mbox{.}(2010)]%
        {lenartz2010enhanced}
\bibfield{author}{\bibinfo{person}{Fabian Lenartz}, \bibinfo{person}{Baptiste
  Mourre}, \bibinfo{person}{Alexander Barth}, \bibinfo{person}{J-M Beckers},
  \bibinfo{person}{Luc Vandenbulcke}, {and} \bibinfo{person}{M Rixen}.}
  \bibinfo{year}{2010}\natexlab{}.
\newblock \showarticletitle{Enhanced ocean temperature forecast skills through
  3-D super-ensemble multi-model fusion}.
\newblock \bibinfo{journal}{\emph{Geophysical Research Letters}}
  \bibinfo{volume}{37}, \bibinfo{number}{19} (\bibinfo{year}{2010}).
\newblock


\bibitem[Letham et~al\mbox{.}(2015)]%
        {letham2015interpretable}
\bibfield{author}{\bibinfo{person}{Benjamin Letham}, \bibinfo{person}{Cynthia
  Rudin}, \bibinfo{person}{Tyler~H McCormick}, {and} \bibinfo{person}{David
  Madigan}.} \bibinfo{year}{2015}\natexlab{}.
\newblock \showarticletitle{Interpretable classifiers using rules and bayesian
  analysis: Building a better stroke prediction model}.
\newblock  (\bibinfo{year}{2015}).
\newblock


\bibitem[Liu et~al\mbox{.}(2003)]%
        {liu2003building}
\bibfield{author}{\bibinfo{person}{Bing Liu}, \bibinfo{person}{Yang Dai},
  \bibinfo{person}{Xiaoli Li}, \bibinfo{person}{Wee~Sun Lee}, {and}
  \bibinfo{person}{Philip~S Yu}.} \bibinfo{year}{2003}\natexlab{}.
\newblock \showarticletitle{Building text classifiers using positive and
  unlabeled examples}. In \bibinfo{booktitle}{\emph{Third IEEE international
  conference on data mining}}. IEEE, \bibinfo{pages}{179--186}.
\newblock


\bibitem[Liu et~al\mbox{.}(2002)]%
        {liu2002partially}
\bibfield{author}{\bibinfo{person}{Bing Liu}, \bibinfo{person}{Wee~Sun Lee},
  \bibinfo{person}{Philip~S Yu}, {and} \bibinfo{person}{Xiaoli Li}.}
  \bibinfo{year}{2002}\natexlab{}.
\newblock \showarticletitle{Partially supervised classification of text
  documents}. In \bibinfo{booktitle}{\emph{ICML}}, Vol.~\bibinfo{volume}{2}.
  Sydney, NSW, \bibinfo{pages}{387--394}.
\newblock


\bibitem[Liu and Wu(2020)]%
        {liu2020fned}
\bibfield{author}{\bibinfo{person}{Yang Liu} {and}
  \bibinfo{person}{Yi-Fang~Brook Wu}.} \bibinfo{year}{2020}\natexlab{}.
\newblock \showarticletitle{Fned: a deep network for fake news early detection
  on social media}.
\newblock \bibinfo{journal}{\emph{ACM Transactions on Information Systems
  (TOIS)}} \bibinfo{volume}{38}, \bibinfo{number}{3} (\bibinfo{year}{2020}),
  \bibinfo{pages}{1--33}.
\newblock


\bibitem[Luo et~al\mbox{.}(2021)]%
        {luo2021pulns}
\bibfield{author}{\bibinfo{person}{Chuan Luo}, \bibinfo{person}{Pu Zhao},
  \bibinfo{person}{Chen Chen}, \bibinfo{person}{Bo Qiao}, \bibinfo{person}{Chao
  Du}, \bibinfo{person}{Hongyu Zhang}, \bibinfo{person}{Wei Wu},
  \bibinfo{person}{Shaowei Cai}, \bibinfo{person}{Bing He},
  \bibinfo{person}{Saravanakumar Rajmohan}, {et~al\mbox{.}}}
  \bibinfo{year}{2021}\natexlab{}.
\newblock \showarticletitle{Pulns: Positive-unlabeled learning with effective
  negative sample selector}. In \bibinfo{booktitle}{\emph{Proceedings of the
  AAAI Conference on Artificial Intelligence}}, Vol.~\bibinfo{volume}{35}.
  \bibinfo{pages}{8784--8792}.
\newblock


\bibitem[Moustafa and Slay(2015)]%
        {moustafa2015unsw}
\bibfield{author}{\bibinfo{person}{Nour Moustafa} {and} \bibinfo{person}{Jill
  Slay}.} \bibinfo{year}{2015}\natexlab{}.
\newblock \showarticletitle{UNSW-NB15: a comprehensive data set for network
  intrusion detection systems (UNSW-NB15 network data set)}. In
  \bibinfo{booktitle}{\emph{2015 military communications and information
  systems conference (MilCIS)}}. IEEE, \bibinfo{pages}{1--6}.
\newblock


\bibitem[Radford et~al\mbox{.}(2019)]%
        {radford2019language}
\bibfield{author}{\bibinfo{person}{Alec Radford}, \bibinfo{person}{Jeffrey Wu},
  \bibinfo{person}{Rewon Child}, \bibinfo{person}{David Luan},
  \bibinfo{person}{Dario Amodei}, \bibinfo{person}{Ilya Sutskever},
  {et~al\mbox{.}}} \bibinfo{year}{2019}\natexlab{}.
\newblock \showarticletitle{Language models are unsupervised multitask
  learners}.
\newblock \bibinfo{journal}{\emph{OpenAI blog}} \bibinfo{volume}{1},
  \bibinfo{number}{8} (\bibinfo{year}{2019}), \bibinfo{pages}{9}.
\newblock


\bibitem[Ribeiro et~al\mbox{.}(2016)]%
        {ribeiro2016should}
\bibfield{author}{\bibinfo{person}{Marco~Tulio Ribeiro},
  \bibinfo{person}{Sameer Singh}, {and} \bibinfo{person}{Carlos Guestrin}.}
  \bibinfo{year}{2016}\natexlab{}.
\newblock \showarticletitle{" Why should i trust you?" Explaining the
  predictions of any classifier}. In \bibinfo{booktitle}{\emph{Proceedings of
  the 22nd ACM SIGKDD international conference on knowledge discovery and data
  mining}}. \bibinfo{pages}{1135--1144}.
\newblock


\bibitem[Sennrich et~al\mbox{.}(2015)]%
        {sennrich2015improving}
\bibfield{author}{\bibinfo{person}{Rico Sennrich}, \bibinfo{person}{Barry
  Haddow}, {and} \bibinfo{person}{Alexandra Birch}.}
  \bibinfo{year}{2015}\natexlab{}.
\newblock \showarticletitle{Improving neural machine translation models with
  monolingual data}.
\newblock \bibinfo{journal}{\emph{arXiv preprint arXiv:1511.06709}}
  (\bibinfo{year}{2015}).
\newblock


\bibitem[Su et~al\mbox{.}(2021)]%
        {su2021positive}
\bibfield{author}{\bibinfo{person}{Guangxin Su}, \bibinfo{person}{Weitong
  Chen}, {and} \bibinfo{person}{Miao Xu}.} \bibinfo{year}{2021}\natexlab{}.
\newblock \showarticletitle{Positive-Unlabeled Learning from Imbalanced Data.}.
  In \bibinfo{booktitle}{\emph{IJCAI}}. \bibinfo{pages}{2995--3001}.
\newblock


\bibitem[Tabik et~al\mbox{.}(2020)]%
        {tabik2020mnist}
\bibfield{author}{\bibinfo{person}{Siham Tabik}, \bibinfo{person}{Ricardo~F
  Alvear-Sandoval}, \bibinfo{person}{Mar{\'\i}a~M Ruiz},
  \bibinfo{person}{Jos{\'e}-Luis Sancho-G{\'o}mez},
  \bibinfo{person}{An{\'\i}bal~R Figueiras-Vidal}, {and}
  \bibinfo{person}{Francisco Herrera}.} \bibinfo{year}{2020}\natexlab{}.
\newblock \showarticletitle{MNIST-NET10: A heterogeneous deep networks fusion
  based on the degree of certainty to reach 0.1\% error rate. Ensembles
  overview and proposal}.
\newblock \bibinfo{journal}{\emph{Information Fusion}}  \bibinfo{volume}{62}
  (\bibinfo{year}{2020}), \bibinfo{pages}{73--80}.
\newblock


\bibitem[Tu et~al\mbox{.}(2022)]%
        {tu2022prevalence}
\bibfield{author}{\bibinfo{person}{Wen-Jun Tu}, \bibinfo{person}{Yimeng Xue},
  {and} \bibinfo{person}{Ding Nie}.} \bibinfo{year}{2022}\natexlab{}.
\newblock \showarticletitle{The prevalence and treatment of diabetes in China
  from 2013 to 2018}.
\newblock \bibinfo{journal}{\emph{JAMA}} \bibinfo{volume}{327},
  \bibinfo{number}{17} (\bibinfo{year}{2022}), \bibinfo{pages}{1706--1706}.
\newblock


\bibitem[Ustun and Rudin(2016)]%
        {ustun2016supersparse}
\bibfield{author}{\bibinfo{person}{Berk Ustun} {and} \bibinfo{person}{Cynthia
  Rudin}.} \bibinfo{year}{2016}\natexlab{}.
\newblock \showarticletitle{Supersparse linear integer models for optimized
  medical scoring systems}.
\newblock \bibinfo{journal}{\emph{Machine Learning}}  \bibinfo{volume}{102}
  (\bibinfo{year}{2016}), \bibinfo{pages}{349--391}.
\newblock


\bibitem[Vanstone et~al\mbox{.}(2013)]%
        {vanstone2013diet}
\bibfield{author}{\bibinfo{person}{Meredith Vanstone}, \bibinfo{person}{Mita
  Giacomini}, \bibinfo{person}{A Smith}, \bibinfo{person}{F Brundisini},
  \bibinfo{person}{Deirdre DeJean}, {and} \bibinfo{person}{Shawn Winsor}.}
  \bibinfo{year}{2013}\natexlab{}.
\newblock \showarticletitle{How diet modification challenges are magnified in
  vulnerable or marginalized people with diabetes and heart disease: a
  systematic review and qualitative meta-synthesis}.
\newblock \bibinfo{journal}{\emph{Ontario health technology assessment series}}
  \bibinfo{volume}{13}, \bibinfo{number}{14} (\bibinfo{year}{2013}),
  \bibinfo{pages}{1}.
\newblock


\bibitem[Wang et~al\mbox{.}(2022)]%
        {wang2022adaptive}
\bibfield{author}{\bibinfo{person}{Yongzhen Wang}, \bibinfo{person}{Jun Lin},
  \bibinfo{person}{Sheng Bi}, \bibinfo{person}{Changlong Sun},
  \bibinfo{person}{Luo Si}, {and} \bibinfo{person}{Xiaozhong Liu}.}
  \bibinfo{year}{2022}\natexlab{}.
\newblock \showarticletitle{Adaptive multi-task positive-unlabeled learning for
  joint prediction of multiple chronic diseases using online shopping
  behaviors}.
\newblock \bibinfo{journal}{\emph{Expert Systems with Applications}}
  \bibinfo{volume}{191} (\bibinfo{year}{2022}), \bibinfo{pages}{116232}.
\newblock


\bibitem[Wei and Li(2018)]%
        {wei2018positive}
\bibfield{author}{\bibinfo{person}{Huihui Wei} {and} \bibinfo{person}{Ming
  Li}.} \bibinfo{year}{2018}\natexlab{}.
\newblock \showarticletitle{Positive and Unlabeled Learning for Detecting
  Software Functional Clones with Adversarial Training.}. In
  \bibinfo{booktitle}{\emph{IJCAI}}. \bibinfo{pages}{2840--2846}.
\newblock


\bibitem[Wilton et~al\mbox{.}(2022)]%
        {wilton2022positive}
\bibfield{author}{\bibinfo{person}{Jonathan Wilton},
  \bibinfo{person}{Abigail~MY Koay}, \bibinfo{person}{Ryan~KL Ko},
  \bibinfo{person}{Miao Xu}, {and} \bibinfo{person}{Nan Ye}.}
  \bibinfo{year}{2022}\natexlab{}.
\newblock \showarticletitle{Positive-Unlabeled Learning using Random Forests
  via Recursive Greedy Risk Minimization}.
\newblock \bibinfo{journal}{\emph{arXiv preprint arXiv:2210.08461}}
  (\bibinfo{year}{2022}).
\newblock


\bibitem[Xu et~al\mbox{.}(2019)]%
        {xu2019revisiting}
\bibfield{author}{\bibinfo{person}{Miao Xu}, \bibinfo{person}{Bingcong Li},
  \bibinfo{person}{Gang Niu}, \bibinfo{person}{Bo Han}, {and}
  \bibinfo{person}{Masashi Sugiyama}.} \bibinfo{year}{2019}\natexlab{}.
\newblock \showarticletitle{Revisiting sample selection approach to
  positive-unlabeled learning: Turning unlabeled data into positive rather than
  negative}.
\newblock \bibinfo{journal}{\emph{arXiv preprint arXiv:1901.10155}}
  (\bibinfo{year}{2019}).
\newblock


\bibitem[Xu et~al\mbox{.}(2013)]%
        {xu2013multi}
\bibfield{author}{\bibinfo{person}{Shuo Xu}, \bibinfo{person}{Xin An},
  \bibinfo{person}{Xiaodong Qiao}, \bibinfo{person}{Lijun Zhu}, {and}
  \bibinfo{person}{Lin Li}.} \bibinfo{year}{2013}\natexlab{}.
\newblock \showarticletitle{Multi-output least-squares support vector
  regression machines}.
\newblock \bibinfo{journal}{\emph{Pattern Recognition Letters}}
  \bibinfo{volume}{34}, \bibinfo{number}{9} (\bibinfo{year}{2013}),
  \bibinfo{pages}{1078--1084}.
\newblock


\bibitem[Yang et~al\mbox{.}(2018)]%
        {yang2018global}
\bibfield{author}{\bibinfo{person}{Chengliang Yang}, \bibinfo{person}{Anand
  Rangarajan}, {and} \bibinfo{person}{Sanjay Ranka}.}
  \bibinfo{year}{2018}\natexlab{}.
\newblock \showarticletitle{Global model interpretation via recursive
  partitioning}. In \bibinfo{booktitle}{\emph{2018 IEEE 20th International
  Conference on High Performance Computing and Communications; IEEE 16th
  International Conference on Smart City; IEEE 4th International Conference on
  Data Science and Systems (HPCC/SmartCity/DSS)}}. IEEE,
  \bibinfo{pages}{1563--1570}.
\newblock


\bibitem[Yu et~al\mbox{.}(2004)]%
        {yu2004pebl}
\bibfield{author}{\bibinfo{person}{Hwanjo Yu}, \bibinfo{person}{Jiawei Han},
  {and} \bibinfo{person}{KC-C Chang}.} \bibinfo{year}{2004}\natexlab{}.
\newblock \showarticletitle{PEBL: Web page classification without negative
  examples}.
\newblock \bibinfo{journal}{\emph{IEEE Transactions on Knowledge and Data
  Engineering}} \bibinfo{volume}{16}, \bibinfo{number}{1}
  (\bibinfo{year}{2004}), \bibinfo{pages}{70--81}.
\newblock


\bibitem[Zhao et~al\mbox{.}(2022)]%
        {zhao2022dist}
\bibfield{author}{\bibinfo{person}{Yunrui Zhao}, \bibinfo{person}{Qianqian Xu},
  \bibinfo{person}{Yangbangyan Jiang}, \bibinfo{person}{Peisong Wen}, {and}
  \bibinfo{person}{Qingming Huang}.} \bibinfo{year}{2022}\natexlab{}.
\newblock \showarticletitle{Dist-PU: Positive-Unlabeled Learning From a Label
  Distribution Perspective}. In \bibinfo{booktitle}{\emph{Proceedings of the
  IEEE/CVF Conference on Computer Vision and Pattern Recognition}}.
  \bibinfo{pages}{14461--14470}.
\newblock


\bibitem[Zhou et~al\mbox{.}(2021)]%
        {zhou2021pure}
\bibfield{author}{\bibinfo{person}{Yao Zhou}, \bibinfo{person}{Jianpeng Xu},
  \bibinfo{person}{Jun Wu}, \bibinfo{person}{Zeinab Taghavi},
  \bibinfo{person}{Evren Korpeoglu}, \bibinfo{person}{Kannan Achan}, {and}
  \bibinfo{person}{Jingrui He}.} \bibinfo{year}{2021}\natexlab{}.
\newblock \showarticletitle{Pure: Positive-unlabeled recommendation with
  generative adversarial network}. In \bibinfo{booktitle}{\emph{Proceedings of
  the 27th ACM SIGKDD Conference on Knowledge Discovery \& Data Mining}}.
  \bibinfo{pages}{2409--2419}.
\newblock


\end{thebibliography}
\end{document}